%% file: main.tex
\documentclass{ifacconf}

\usepackage{graphicx}      
\usepackage{natbib}        
\pdfoutput=1

\usepackage{amsmath}
\usepackage{amsfonts}
\usepackage{dsfont}
\usepackage{subcaption}
\usepackage{tikz}
\usepackage{pgf}
\usetikzlibrary{plotmarks}
\usetikzlibrary{arrows}
\usepackage{pgfplots}
\newlength\figureheight
\newlength\figurewidth

\newtheorem{remark}{Remark}

\usepackage{algpseudocode}
\usepackage{algorithm}

\DeclareCaptionLabelFormat{TABLEhax}{\normalsize{Table}}

\newif\ifcomment
\commenttrue
\ifcomment
\newcommand{\comments}[1]{\textcolor{blue}{#1}}
\else
\newcommand{\comments}[1]{}
\fi

\DeclareMathOperator{\diag}{diag}

\newcommand{\fref}[1]{Figure~\ref{#1}}

\newcommand{\eref}[1]{(\ref{#1})}
\newcommand{\sref}[1]{Section~\ref{#1}}

\begin{document}
\begin{frontmatter}

\title{Evaluation of Spectral Learning for the Identification of Hidden Markov Models} 

\thanks[footnoteinfo]{This work was partially supported by the Swedish Research
Council and the Linnaeus Center ACCESS at KTH. The research leading to these
results has received funding from The European Research Council under the
European Community's Seventh Framework program (FP7 2007-2013) / ERC Grant
Agreement N. 267381.}

\author[First]{Robert Mattila} 
\author[First]{Cristian R. Rojas} 
\author[First]{Bo Wahlberg}

\address[First]{Department of Automatic Control and ACCESS, School of
Electrical Engineering, KTH Royal Institute of Technology, SE-100 44 Stockholm,
Sweden. (e-mails: rmattila@kth.se, \{cristian.rojas, bo.wahlberg\}@ee.kth.se).}

\begin{abstract}                
    Hidden Markov models have successfully been applied as models of discrete
time series in many fields. Often, when applied in practice, the parameters of
these models have to be estimated. The currently predominating identification
methods, such as maximum-likelihood estimation and especially
expectation-maximization, are iterative and prone to have problems with local
minima. A non-iterative method employing a spectral subspace-like approach has
recently been proposed in the machine learning literature. This paper evaluates
the performance of this algorithm, and compares it to the performance of the
expectation-maximization algorithm, on a number of numerical examples. We find
that the performance is mixed; it successfully identifies some systems with
relatively few available observations, but fails completely for some systems even
when a large amount of observations is available. An open question is how this
discrepancy can be explained. We provide some indications that it could be
related to how well-conditioned some system parameters are.

\end{abstract}

\begin{keyword}
    spectral learning, hidden Markov models, HMM, system identification,
    spectral factorization, method of moments, performance evaluation 
\end{keyword}

\end{frontmatter}

\section{Introduction}

\emph{Hidden Markov Models} (HMMs) are standard tools for modeling discrete
time series. They have, among others, been applied to such disperse fields as
speech recognition (e.g. \cite{rabiner:speech}, \cite{gales:speech}), genomic
sequence analysis (e.g.  \cite{eddy:dna_hmm}) and financial stock prediction
(e.g.  \cite{hassan:stock}). The parameters of these models are usually unknown
and have to be estimated from data when applied in practice. Methods for
performing this task, such as the \emph{Expectation-Maximization} (EM)
algorithm (also referred to as the Baum-Welch algorithm when applied
specifically to HMMs), are usually iterative and might encounter problems with
local minima and slow convergence (see e.g. \cite{hsu:hmm}).

A well-known approach for the identification of linear states-space models is
to employ subspace techniques (e.g. \cite{van:subspace}, or
\cite{ljung:subspace}). This approach is non-iterative and avoids the concern
of local minima that can cause problems when employing EM. There have been
attempts to apply similar methods to HMMs (e.g.  \cite{vanluyten:approach},
\cite{hjalmarsson:4s}, \cite{anand:mom}, and \cite{hsu:hmm}). One of the
difficulties with adapting these techniques is that the HMM has some
restrictions that are not present for regular linear systems, namely that some
parameters represent probabilities and thus have to be non-negative and sum to
one.

This paper is concerned with the so-called \emph{spectral learning} algorithm
proposed by \citet{hsu:hmm}. The idea of their method is to relate observable
quantities, correlations in pairs and triplets in a sequence of observations,
to the system parameters. The observable quantities are estimated from the
available data and used to recover estimates of the parameters of the HMM by
reversing the relations. This is fundamentally a method of moments. It has been
generalized to other models and put in a tensor decomposition framework in
subsequent work, see e.g. \cite{anandkumar:tensor}, \cite{anand:mom}, and
\cite{anandkumar:rank1tensor}.

The contribution of this paper is an evaluation of how the spectral learning
algorithm performs on a number of systems. We find that its performance is
varied. It identifies some systems well using a small amount of
observations, but fails to identify some systems even with a large number of
observations.

This paper is structured as follows: The HMM is formally introduced in
\sref{sec:preliminaries}, along with statistical moments. \sref{sec:problem}
states the identification problem for HMMs. \sref{sec:SL} outlines
the spectral learning algorithm by \cite{hsu:hmm}. \sref{sec:error} discusses how the
correctness of the estimates is measured. \sref{sec:numerical} provides
numerical results from simulations for SL and EM. \sref{sec:conclusion}
concludes the paper with a brief discussion of the results.

\section{Preliminaries}
\label{sec:preliminaries}

Vectors in this paper are assumed to be column vectors and we use the symbol
$\; \widehat\; \;$ to denote an empirical estimate of a numerical quantity.

\subsection{Hidden Markov Models}

The HMM is a generalization of the Markov chain, which is a discrete stochastic
model that can be represented as a sequence of stochastic variables, say $x_1,
x_2, \dots$ These variables take values in some set $\mathcal{X} = \{1, 2,
\dots, X\}$, which is called the state-space (of dimension $X \in
\mathbb{N}^+$). The crucial assumption of a Markov chain is that the Markov
property,
\begin{align} 
        \Pr[ x_{k+1} = i &| x_1 = i_1, x_2 = i_2, \dots, x_k = i_k] \notag \\
        &= \Pr[x_{k+1} = i | x_k = i_k],
\end{align}
holds.

The quantity $\Pr[x_{k+1} = i | x_k = j]$ is called a \emph{transition
probability}. If these quantities do not depend on absolute time, then
it is possible to summarize all transition probabilities in a constant matrix
$T \in [0,1]^{X \times X}$, known as the \emph{transition matrix}, with elements
\begin{equation}
    [T]_{ij} = \Pr[x_{k+1} = i | x_k = j].
\end{equation}
This matrix is \emph{column stochastic}, i.e., it has to satisfy elementwise
non-negativity,
\begin{equation}
    [T]_{ij} \in [0, 1] \; \forall \; i,j = 0, 1, \dots, X,
\end{equation}
and the elements in each column must sum to one,
\begin{equation}
    \sum_{i = 1}^X [T]_{ij} = 1 \; \forall \; j = 0, 1, \dots, X.
\end{equation}

The state of the Markov chain is not directly observed in an HMM, but rather an
observation $y_k$ is made out of a finite set $\mathcal{Y} = \{1, 2, \dots,
Y\}$ at each time instant $k$, where $Y \in \mathbb{N}^+$ is the number of
possible observations. The observation made at time $k$ is a stochastic variable, but is
related to the (hidden) state of the Markov chain. The \emph{observation
probabilities} of an HMM can be written as a matrix $O \in [0,1]^{Y \times X}$,
referred to as the \emph{observation matrix}, with elements:
\begin{equation}
    [O]_{ij} = \Pr[y_k = i | x_k = j].
\end{equation}
This matrix is also column stochastic and satisfies similar relations as those
for $T$.

These two quantities along with the initial distribution $\pi_0 \in [0,1]^X$ of
the Markov chain, defined as
\begin{equation}
    [\pi_0]_i = \Pr[ x_1 = i ],
\end{equation}
completely specify the HMM. A more complete description of the HMM can be found
in \cite{ryden:hmm}.

\subsection{Moments}

In the spectral learning algorithm, the joint probabilities of $n$-tuples of
consecutive observations are decomposed into the parameters of the HMM. These
are called $n$th order moments. Only the first, second and third order moments
are utilized, which are vector, matrix and third order tensor quantities,
respectively, and which are defined as follows:
\begin{equation}
    [S_1]_i = \Pr[y_1 = i],
\end{equation}
for $i = 1, 2, \dots, Y$, and
\begin{equation}
    [S_{2, 1}]_{ij} = \Pr[y_2 = i, y_1 = j],
\end{equation}
for $i, j = 1, 2, \dots, Y$. The third order moments can be written as $Y$
matrices if one index is fixed:
\begin{equation}
    [S_{3, y, 1}]_{ij} = \Pr[y_3 = i, y_2 = y, y_1 = j],
\end{equation}
for $i, j, y = 1, 2, \dots, Y$. Here, $y_1$ refers to the first item in
a triplet of consecutive observations, $y_2$ to the second and $y_3$ to the
third.

Note that it is possible to define other moments of the same orders, for
example $S_{3,1}$, in an analogous fashion.

These quantities can straightforwardly be estimated from data by sampling
triplets of observations from the HMM and calculating the relative frequency of
different singletons, pairs and triplets.  If a long consecutive sequence of
observations is available, then either a sliding window or an independent
sampling approach can be used to generate triplets.

\section{Problem Formulation}
\label{sec:problem}

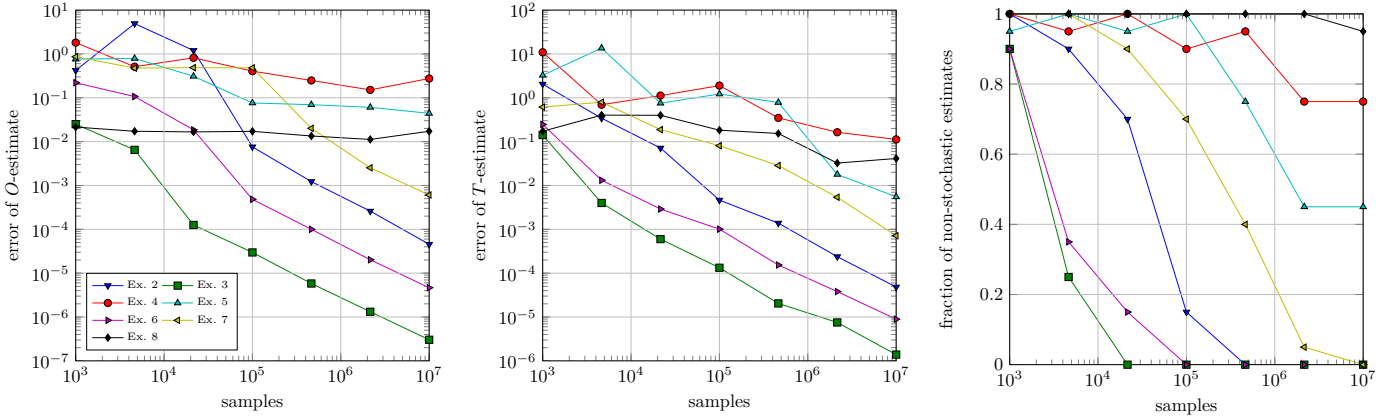
\begin{figure*}[t!]
\begin{subfigure}[h]{0.33333\textwidth}
        \centering

        \setlength\figureheight{8.4cm} 
        \setlength\figurewidth{8.4cm} 

        \resizebox{\linewidth}{!}{
        \input{./figures/chap5_SL_bench/O_dev.tikz}}

    \end{subfigure}
    \begin{subfigure}[h]{0.33333\textwidth}
        \centering

        \setlength\figureheight{8.4cm} 
        \setlength\figurewidth{8.4cm} 

        \resizebox{\linewidth}{!}{
        \input{./figures/chap5_SL_bench/T_dev.tikz}}

    \end{subfigure}
    \begin{subfigure}[h]{0.33333\textwidth}
        \centering

        \setlength\figureheight{8.4cm} 
        \setlength\figurewidth{8.4cm} 

        \resizebox{\linewidth}{!}{
        \input{./figures/chap5_SL_bench/non_stoch.tikz}}

    \end{subfigure}

    \caption{Performance of SL on seven systems. All points in the plots are
        averaged over 20 simulations. The two leftmost plots show the MSE of
        the estimated observation ($O$) and transition ($T$) matrices. The
        rightmost plot shows the fraction of the 20 simulations that resulted
        in some element of $\hat T$ or $\hat O$ being negative or having a non-zero
    imaginary component.}

    \label{fig:SL_bench}
\end{figure*}

The problem we aim to solve is the following: At hand is an HMM-system
with unknown parameters. The data that we have available
is a sequence of consecutive observations from the system. To be able to
construct for example an HMM-filter, we want to fit parameters of an HMM such
that it models the observed data well. We assume that we know the number of
hidden states $X$ and the number of possible observations $Y$ of the true
system.


\section{The Spectral Learning Algorithm}
\label{sec:SL}

\citet{hsu:hmm} are not mainly concerned with recovering explicit expressions
for estimates of the parameters $T$, $O$ and $\pi_0$, but rather intend to
identify another parametrization of the HMM. This parametrization allows them
to calculate the probabilities of sequences of observations, for example, the
joint distribution $\Pr[y_1, y_2, \dots, y_k]$ and the conditional distribution
$\Pr[y_{k+1} | y_1, y_2, \dots, y_{k}]$. They describe in an appendix how the
method by \citet{mossel:markov} can be used in conjunction with their method to
recover explicit expressions for $\hat T$, $\hat O$ and $\hat \pi_0$. This
combined method will be outlined below, and is what we refer to as
\emph{Spectral Learning} (SL).

It is possible to derive the following relations between different moments and
the parameters of the HMM:
\begin{align}
    S_1 &= O \pi_0, \\
    S_{2,1} &= OT \diag( \pi_0 ) O^T, \\
    S_{3,1} &= OT^2 \diag( \pi_0 ) O^T
\end{align}
and
\begin{equation}
    S_{3,y,1} = OT \diag(e_y^TO)T\diag(\pi_0)O^T,
\end{equation}
where $e_i$ is the $i$th canonical (column) unit vector. Details can be found
in e.g. \citet{hsu:hmm}, \citet{mattila:msc} and \citet{johnson:simple}.

\citet{hsu:hmm} make the assumptions that $\pi_0 > 0$ elementwise and that $T$
and $O$ have rank $X$. To be able to handle the case that $Y > X$, a matrix $U$
of the left singular values of $S_{2,1}$ corresponding to the $X$ largest
singular values is introduced. 

\begin{algorithm}
    \caption{Spectral Learning from \cite{hsu:hmm}}

  \begin{algorithmic}[1]
      \State Sample triplets of observations $(y_1, y_2, y_3)$ from the HMM
      and form empirical estimates $\hat S_{1}$, $\hat S_{2,1}$, $\hat S_{3,1}$
      and $\hat S_{3,y,1}$ for $y = 1, \dots, Y$.
      \State Calculate the SVD of $\hat S_{2,1}$ and form $\hat U$ by taking
      the left singular vectors corresponding to the $X$ largest singular
      values as columns.
      \State Generate $Y$ normally distributed random variables $\{g_y \sim N(0,1) : y = 1, \dots,
      Y \}$.
      \State Perform an eigendecomposition of the left hand side of
      \eref{eq:SL_rand_eig} when the above estimates are used, i.e. of
            \begin{equation}
                \sum_{y = 1}^Y g_y (\hat U^T \hat S_{3, y, 1})(\hat
U^T \hat S_{3,1})^+.
            \notag
            \end{equation}
        \State Use the matrix of eigenvectors from Step 4 to diagonalize
        $(\hat U^T \hat S_{3,y,1})(U^T \hat S_{3,1})^+$ for $y = 1, \dots, Y$ and
        take the diagonal as row $y$ of $\hat O$. The diagonalizations are
        performed by multiplicating from the left and by the inverse from the
        right in accordance with \eref{eq:SL_diag}.
        \State Calculate 
            \begin{equation}
                \hat \pi \leftarrow \hat O^+ \hat S_1
                \notag
            \end{equation}
            and
            \begin{equation}
                \hat T \leftarrow \hat O^+ \hat S_{2,1} (\hat O^+)^T \diag(\hat
\pi_0)^{-1}.
                \notag
            \end{equation} 
        \State Return $\hat O$, $\hat T$ and $\hat \pi_0$.
  \end{algorithmic}
    \label{alg:SL}
\end{algorithm}

By employing the relations given above, it can be shown that the following
expression holds for $y = 1, 2, \dots, Y$:
\begin{equation}
    (U^T S_{3, y, 1})(U^T S_{3, 1})^+ =
    (U^T OT)\diag(e_y^T O)(U^T OT)^{-1},
    \label{eq:SL_diag_first}
\end{equation}
where $^+$ denotes the Moore-Penrose pseudo-inverse. 

What should be noted here is that everything on the left hand side of
\eref{eq:SL_diag_first} can be estimated from data and that the right hand side
is an eigenfactorization where the diagonal matrix contains the $y$th row of the
observation matrix.

The idea of SL is to recover one row of $O$ at a time using the trivial
reformulation of \eref{eq:SL_diag_first}:
\begin{equation}
   (U^T OT)^{-1}(U^T S_{3, y, 1})(U^T S_{3, 1})^+(U^T OT) =
    \diag(e_y^T O). 
    \label{eq:SL_diag}
\end{equation}
Notice that the same transformation $(U^TOT)$ diagonalizes
\eref{eq:SL_diag_first} for every $y$. This is important for preserving the
order of the eigenvalues.

The transformation could be recovered for any $y$,
but to increase the robustness of the method, a set of random variables,
$g_i \sim \mathcal{N}(0, 1)$ for $i = 1, 2, \dots, Y$, is introduced. Here,
$\mathcal{N}(\mu, \sigma^2)$ is the normal distribution with mean value $\mu$
and variance $\sigma^2$.
It can be shown from \eref{eq:SL_diag_first} that 
\begin{align}
    \sum_{y = 1}^Y g_y (U^T S_{3, y, 1})&(U^T S_{3,1})^+  = \notag \\
     (U^TOT) \Big\{ & \sum_{y = 1}^Y g_y \diag( e_y^T O ) \Big\}
(U^TOT)^{-1}
    \label{eq:SL_rand_eig}
\end{align}
holds.

Thus, by performing an eigendecomposition of the left hand side of
\eref{eq:SL_rand_eig}, we recover the diagonalization transformation $(U^TOT)$,
up to scaling and permutation of the columns. This matrix is then utilized in
\eref{eq:SL_diag} to recover each row of $O$.

\begin{remark}
If for each $y$ an eigendecomposition of \eref{eq:SL_diag_first} is performed 
directly using some numerical method, then the elements of $O$ will be recovered,
but since eigenvalues do not have any intrinsic ordering, as $O$ is assembled
(row by row), every row will have its elements sorted according to the order
in which the eigenvalues were returned by the method performing the
eigendecomposition. Since the columns of $O$ correspond to the different hidden
states of the HMM, a consistent ordering is required. By using the same
diagonalization matrix --- the one recovered in \eref{eq:SL_rand_eig} --- we
guarantee that the eigenvalues have a consistent ordering.
\end{remark}

Once $O$ is recovered, \citet{hsu:hmm} suggest that the following expressions
be used to recover the other parameters:
\begin{equation}
    \pi_0 = O^+ S_1
\end{equation}
and
\begin{equation}
    T = O^+ S_{2,1} (O^+)^T \diag( \pi_0 )^{-1}.
\end{equation}

To employ the method on real data, estimates are to be used for the moments
$S_1$, $S_{2,1}$, $S_{3,1}$ and $S_{3,y,1}$. This concludes the SL algorithm,
which for convenience is summarized in Algorithm~\ref{alg:SL}.

\begin{remark} Since in our problem formulation we make the assumption that all
    our observations are consecutive, unlike \citet{hsu:hmm} who independently
    sample triplets of observations to estimate the different moments, our
    estimate of the initial distribution will be an estimate of
    the stationary distribution of the Markov chain corresponding to the
    transition matrix $T$. 
\end{remark}

\section{Error Measure}
\label{sec:error}

We use the \emph{Mean Squared Error} (MSE) of the elements in the relevant
matrices as a measure of the error of the estimates. For two matrices $A, \hat
A \in \mathbb{R}^{m\times n}$, the MSE is defined as
\begin{equation}
    \text{MSE}(A, \hat A) = \frac{ \| A - \hat A \|_F^2 }{m \times n},
\end{equation}
where $\| \cdot \|_F$ denotes the Frobenius norm.

It is important to notice that the columns of $\hat O$, and the rows and columns
of $\hat T$, can be permuted compared to those in $O$ and $T$, depending on the
method used to calculate the diagonalization transformation of
\eref{eq:SL_rand_eig}. It is thus not meaningful to calculate and present, for example,
$\text{MSE}(O, \hat O)$ directly.

The columns of $\hat O$, and the rows and columns of $\hat T$, have to be
permuted as to line up with those of the original matrices used to generate the
sequence of observations if $\text{MSE}(\cdot, \hat \cdot)$ is to make sense.
This is a simple combinatorial problem which is solved in the results to be
presented. The intuition behind this is that the order of the \emph{hidden}
states can not be discerned from external observations of the HMM (nor in the
case of a regular linear system). It follows from the SL algorithm that HMMs
are generically identifiable up to these permutations.

Also notice that there is nothing in SL that enforces the stochasticity
constraints on the estimates of $T$, $O$ and $\pi_0$. The elements of $\hat O$
are the result of an eigendecomposition, and can therefore turn out to be
negative, or even complex, numbers. The elements of $\hat T$ and $\hat \pi_0$
might also lie outside of $[0,1]$, since they are calculated from
$\hat O$. This is currently one of the main difficulties of using SL.

\section{Numerical Results}
\label{sec:numerical}

This section presents numerical results for SL and EM. All simulations were
performed on a 1.3 GHz MacBook Air with 4 GB RAM.

\subsection{Spectral Learning}

The performance of SL\footnote{The implementation and simulations were
performed using \mbox{MATLAB} R2013a.} was evaluated on seven systems (taken
from standard texts on HMMs or conceived) and the
results are presented in Figures~\ref{fig:SL_bench} and \ref{fig:time}. The
systems had three hidden states, and three (Examples 2, 4, 5, 6 and 7) or ten
(Examples 3 and 8) possible observations, i.e.  $X = 3$ and $Y \in \{3, 10\}$.
The exact numerical parameters (i.e. the transition and observation matrices)
for each example can be found in \cite{mattila:msc}. Every point in the plots
is the average of 20 simulations.  

As previously mentioned, there is no guarantee that the estimates recovered by
SL are (stochastically) valid. In the rightmost plot of
Figure~\ref{fig:SL_bench}, the fraction of these 20 simulations that
resulted in estimates with elements with negative or non-zero imaginary
components is presented. 

SL demonstrated a mixed performance on the examples on which it was evaluated. It
converged for a relatively small amount of samples on Examples 3 and 6, but
failed to converge even for large amounts of samples for Examples 4, 5 and 8.
This can be seen both from the error in the estimates of $T$ and $O$, but also
from the fraction of invalid estimates. The performance on Examples 2 and
7 was in between these other two performance classes.

It is worth noting that the slopes of the errors appear to be constant once SL
starts providing a large fraction of valid estimates, which happened for
different amounts of samples for the different systems.

\subsection{Expectation-Maximization Method}

\begin{figure}[]
    \begin{subfigure}[h]{0.24\textwidth}
        \centering $\quad$\tiny{\textbf{SL}}
    \end{subfigure}
    \begin{subfigure}[h]{0.24\textwidth}
        \centering $\quad$\tiny{\textbf{EM}}
    \end{subfigure}
    \begin{subfigure}[h]{0.24\textwidth}
        \centering

        \setlength\figureheight{8.4cm} 
        \setlength\figurewidth{8.4cm} 

        \resizebox{\linewidth}{!}{
        \input{./figures/chap5_SL_bench/time_avg.tikz}}

    \end{subfigure}
    \begin{subfigure}[h]{0.24\textwidth}
        \centering

        \setlength\figureheight{8.4cm} 
        \setlength\figurewidth{8.4cm} 

        \resizebox{\linewidth}{!}{
        \input{./figures/chap5_EM/time_avg.tikz}}

    \end{subfigure}

    \caption{Computational time, as measured from the point that data is given to
        the algorithm until estimates of the parameters are returned, for SL
        and EM on the different systems.  Points in the plots are averaged over
        3 simulations for EM and 20 for SL. Notice the scales of the time-axes
    in the two plots. }
    \label{fig:time}
\end{figure}
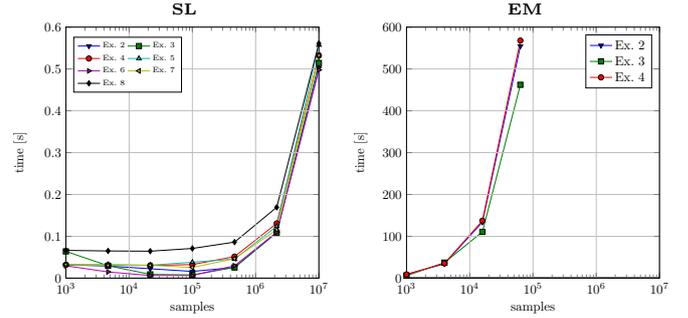

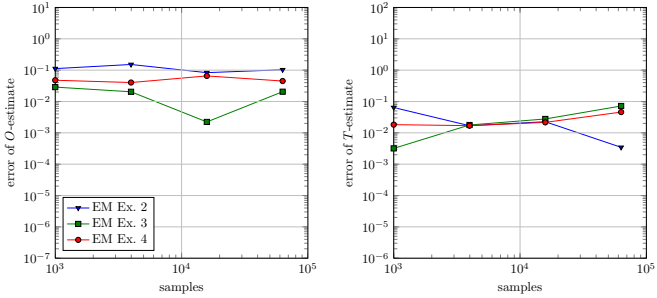
\begin{figure}[t!]
    \begin{subfigure}[h]{0.49\columnwidth}
        \centering

        \setlength\figureheight{8.4cm} 
        \setlength\figurewidth{8.4cm} 

        \resizebox{\linewidth}{!}{
        \input{./figures/chap5_EM/O_dev.tikz}}

    \end{subfigure}
    \begin{subfigure}[h]{0.49\columnwidth}
        \centering

        \setlength\figureheight{8.4cm} 
        \setlength\figurewidth{8.4cm} 

        \resizebox{\linewidth}{!}{
        \input{./figures/chap5_EM/T_dev.tikz}}

    \end{subfigure}

    \caption{MSE of the estimates of the observation ($O$) and transition ($T$)
matrices for the EM-algorithm on three different systems when random matrices
were used as initial guesses.}
    \label{fig:EM}
\end{figure}

For a thorough treatment of the EM-algorithm, see \cite{krishnan:em}. The
implementation provided in MATLAB (\texttt{hmmtrain}) was used with the maximum
number of iterations taken as the default value of 500.
Three of the examples on which SL had varying performance were considered. In
Figures~\ref{fig:time} and \ref{fig:EM}, the EM-algorithm was started with
random matrices as initial guesses for the HMM parameters. 

 EM performed better, errorwise, than SL on Examples 2 and 4 when a small
 amount of observations was available. However, as is apparent in
 Figure~\ref{fig:time}, the time-scale is orders of magnitude larger than that
 of SL. For all of the examples that were considered, when about $10^5$
 available samples were available, SL delivered estimates in less than
 a tenth of a second, whereas EM required about ten minutes.

 The initial guesses for the parameters in EM play an important role as can be seen in
 Figure~\ref{fig:EM}; the error of the estimated $T$ matrix increased for two of the
 systems as more and more of samples were available. This is probably due to
the random initial guesses being worse in those simulations.

\begin{figure}[b!]
    \begin{subfigure}[h]{0.49\columnwidth}
        \centering

        \setlength\figureheight{8.4cm} 
        \setlength\figurewidth{8.4cm} 

        \resizebox{\linewidth}{!}{
        \input{./figures/EM_with_true_and_SL/O_dev.tikz}}

    \end{subfigure}
    \begin{subfigure}[h]{0.49\columnwidth}
        \centering

        \setlength\figureheight{8.4cm} 
        \setlength\figurewidth{8.4cm} 

        \resizebox{\linewidth}{!}{
        \input{./figures/EM_with_true_and_SL/T_dev.tikz}}

    \end{subfigure}

    \caption{MSE of the estimates of the observation ($O$) and transition ($T$)
matrices for the EM-algorithm on three different systems when the true matrices
were used as initial guesses (solid), along with the corresponding results of
SL from \fref{fig:SL_bench} (dashed). }
    \label{fig:EM_TRUE}
\end{figure}
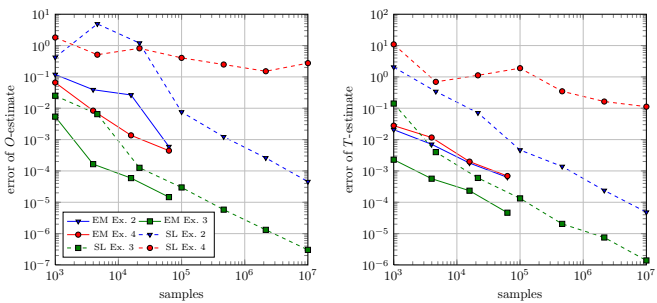

 In Figure~\ref{fig:EM_TRUE}, the true parameters of the HMMs were used as initial
 guesses in the EM algorithm. To make a comparison between EM and SL easier, we
 also plot the performance of SL (from \fref{fig:SL_bench}) as dashed lines in
 the same figure. 
 
 The error for EM when started at the true system parameters is closely related
 to the Cram\'{e}r-Rao bounds for how well any estimation method can perform.
 SL was close to the performance of EM for Example 3, but was far off
 for the other two examples. Note that the EM curves lie higher for those
 two examples than for Example 3, implying that the maximum of the likelihood
 function for the available data lies further away from the true maximum.

\subsection{Relation to Condition Numbers}


The mixed performance of SL, compared both to EM and to itself on the different
examples, could be due to many reasons. One could be that there is a loss of
information when only the first, second and third order moments are used,
instead of moments up to the length of the whole observation sequence.

Regarding the discrepancy between the different examples, a simple
perturbation analysis (\cite{mattila:msc}) shows that the condition number of
$(UOT)$ is related to the upper bound on how far the eigenvalues might move
in \eref{eq:SL_diag_first} (i.e. the elements of one row of the recovered
observation matrix) due to perturbations in the estimates of the moments. 

Since the matrix $U$ in SL is not absolutely specified (the choice as some of
the singular vectors is just a suggestion, see \cite{hsu:hmm} for details), we
provide in Table~1 the condition number of $OT$ for the different systems in
\fref{fig:SL_bench}. We have sorted the examples after what the MSE was when
$10^7$ samples were available in the estimation procedure. The condition
numbers are perfectly sorted except for the outlier Example 8. This
suggests that the condition number of $OT$ could give some indication on how
well SL will perform for a given system. 

\section{Conclusion}
\label{sec:conclusion}

We have implemented and evaluated the algorithm outlined in \citet{hsu:hmm},
which uses spectral methods to deliver one-shot estimates of the parameters of
an HMM. We saw that the performance of the algorithm is overall good, but that
it fails completely for some systems, and that the performance could be related
to a certain condition number. 

We also observed that SL is orders of magnitude faster than EM, especially when
a large amount of observations is available. This suggests that SL could be
a preferable alternative to EM in cases where there is a large number of observations
available from the HMM. The recovered estimates can then be refined, if
necessary, using for example EM. Note that this assumes that the estimates
recovered from SL are stochastically valid, which was not the case for many
systems. SL started providing valid estimates when different amounts of samples
were available for the different systems. 

\begin{table}[t!]
    \captionsetup[sub]{labelformat=TABLEhax}
    \subcaption{\normalsize{1. The examples from \fref{fig:SL_bench} sorted
according to the error in the estimate of the $O$-matrix when $10^7$ samples
were used, together with the condition number of the product $OT$.}}
    \centering
    \begin{tabular}{| c | c | c | }
        \hline
        Ex. & Error of $O$-estimate with $10^7$ samples & cond($OT$) \\ 
        \hline
            4 & 2.75$\times 10^{-1}$ & 115.4 \\ 
            5 & 4.42$\times 10^{-2}$ & 27.1 \\ 
            8 & 1.72$\times 10^{-2}$ & 208.3 \\ 
            7 & 6.06$\times 10^{-4}$ & 21.6 \\ 
            2 & 4.55$\times 10^{-5}$ & 10.8 \\ 
            6 & 4.66$\times 10^{-6}$ & 5.4 \\ 
            3 & 3.03$\times 10^{-7}$ & 2.6 \\
        \hline
    \end{tabular}
    \label{table:condition}
\end{table}

\begin{ack}
The authors would like to thank Vikram Krishnamurthy for helpful discussions. 
\end{ack}

\addtolength{\textheight}{-14cm}

\bibliography{references}

\end{document}

%% file: figures/chap5_SL_bench/O_dev.tikz
%
%
%
%
\begin{tikzpicture}

\definecolor{color1}{rgb}{0.75,0,0.75}
\definecolor{color0}{rgb}{0,0.75,0.75}
\definecolor{color2}{rgb}{0.75,0.75,0}

\begin{loglogaxis}[
xlabel={samples},
ylabel={error of $O$-estimate},
xmin=1000, xmax=10000000,
ymin=1e-07, ymax=10,
axis on top,
width=\figurewidth,
height=\figureheight,
xmajorgrids,
ymajorgrids,
legend columns=2,
legend style={at={(0.03,0.03)}, anchor=south west, font=\tiny},
legend entries={{Ex. 2},{Ex. 3},{Ex. 4},{Ex. 5},{Ex. 6},{Ex. 7},{Ex. 8}}
]
\addplot [very thin, blue, mark=triangle*, mark size=2, mark options={rotate=180,draw=black}]
coordinates {
(1000,0.42068676)
(4641,4.9121650574)
(21544,1.1980582023)
(100000,0.0075926634)
(464158,0.0012276053)
(2154434,0.0002602418)
(10000000,4.55481e-05)

};
\addplot [very thin, green!50.0!black, mark=square*, mark size=2, mark options={draw=black}]
coordinates {
(1000,0.0250202536)
(4641,0.0064876399)
(21544,0.0001257871)
(100000,2.96942e-05)
(464158,5.8149e-06)
(2154434,1.3195e-06)
(10000000,3.03e-07)

};
\addplot [very thin, red, mark=*, mark size=2, mark options={draw=black}]
coordinates {
(1000,1.8094939353)
(4641,0.5112988391)
(21544,0.8147439593)
(100000,0.4045767371)
(464158,0.2482228942)
(2154434,0.151350762)
(10000000,0.2749241359)

};
\addplot [very thin, color0, mark=triangle*, mark size=2, mark options={draw=black}]
coordinates {
(1000,0.7618054034)
(4641,0.7894723745)
(21544,0.3096399136)
(100000,0.0761646922)
(464158,0.0695562198)
(2154434,0.0605903666)
(10000000,0.0441909001)

};
\addplot [very thin, color1, mark=triangle*, mark size=2, mark options={rotate=270,draw=black}]
coordinates {
(1000,0.2202443175)
(4641,0.1066200207)
(21544,0.0185473436)
(100000,0.0004833558)
(464158,0.0001000573)
(2154434,2.02093e-05)
(10000000,4.661e-06)

};
\addplot [very thin, color2, mark=triangle*, mark size=2, mark options={rotate=90,draw=black}]
coordinates {
(1000,0.8638069555)
(4641,0.4753318388)
(21544,0.4864641529)
(100000,0.4809091875)
(464158,0.0201087986)
(2154434,0.002529429)
(10000000,0.0006058439)

};
\addplot [very thin, black, mark=diamond*, mark size=2]
coordinates {
(1000,0.0214877946)
(4641,0.0172289103)
(21544,0.0166569148)
(100000,0.0171562518)
(464158,0.0134477027)
(2154434,0.0112140707)
(10000000,0.017234292)

};

\end{loglogaxis}

\end{tikzpicture}

%% file: figures/chap5_SL_bench/T_dev.tikz
%
%
%
%
\begin{tikzpicture}

\definecolor{color1}{rgb}{0.75,0,0.75}
\definecolor{color0}{rgb}{0,0.75,0.75}
\definecolor{color2}{rgb}{0.75,0.75,0}

\begin{loglogaxis}[
xlabel={samples},
ylabel={error of $T$-estimate},
xmin=1000, xmax=10000000,
ymin=1e-06, ymax=100,
axis on top,
width=\figurewidth,
height=\figureheight,
xmajorgrids,
ymajorgrids
]
\addplot [very thin, blue, mark=triangle*, mark size=2, mark options={rotate=180,draw=black}]
coordinates {
(1000,2.044555523)
(4641,0.3422954473)
(21544,0.0713145182)
(100000,0.0046401567)
(464158,0.0013853535)
(2154434,0.0002382684)
(10000000,4.828e-05)

};
\addplot [very thin, green!50.0!black, mark=square*, mark size=2, mark options={draw=black}]
coordinates {
(1000,0.1412195755)
(4641,0.0039895532)
(21544,0.0005980195)
(100000,0.0001330955)
(464158,2.04241e-05)
(2154434,7.5226e-06)
(10000000,1.3918e-06)

};
\addplot [very thin, red, mark=*, mark size=2, mark options={draw=black}]
coordinates {
(1000,10.9115376712)
(4641,0.6947624578)
(21544,1.121573524)
(100000,1.8971174293)
(464158,0.3474508554)
(2154434,0.1636030512)
(10000000,0.1122621116)

};
\addplot [very thin, color0, mark=triangle*, mark size=2, mark options={draw=black}]
coordinates {
(1000,3.2647493832)
(4641,13.5495381832)
(21544,0.7516025176)
(100000,1.2214622034)
(464158,0.7820289939)
(2154434,0.0178942815)
(10000000,0.0055125861)

};
\addplot [very thin, color1, mark=triangle*, mark size=2, mark options={rotate=270,draw=black}]
coordinates {
(1000,0.2474697269)
(4641,0.0131670055)
(21544,0.0028995393)
(100000,0.0010088417)
(464158,0.0001536687)
(2154434,3.79999e-05)
(10000000,8.9309e-06)

};
\addplot [very thin, color2, mark=triangle*, mark size=2, mark options={rotate=90,draw=black}]
coordinates {
(1000,0.6135164961)
(4641,0.7932072783)
(21544,0.1877241847)
(100000,0.0807379305)
(464158,0.0285079205)
(2154434,0.0053860512)
(10000000,0.0007133789)

};
\addplot [very thin, black, mark=diamond*, mark size=2]
coordinates {
(1000,0.1724298704)
(4641,0.4018634417)
(21544,0.3987543084)
(100000,0.1822255021)
(464158,0.1533024798)
(2154434,0.0323905493)
(10000000,0.04130079)

};

\end{loglogaxis}

\end{tikzpicture}

%% file: figures/chap5_SL_bench/non_stoch.tikz
%
%
%
%
\begin{tikzpicture}

\definecolor{color1}{rgb}{0.75,0,0.75}
\definecolor{color0}{rgb}{0,0.75,0.75}
\definecolor{color2}{rgb}{0.75,0.75,0}

\begin{semilogxaxis}[
xlabel={samples},
ylabel style={align=center, text width=7.4cm},
ylabel={fraction of non-stochastic estimates},
xmin=1000, xmax=10000000,
ymin=0, ymax=1,
axis on top,
width=\figurewidth,
height=\figureheight,
xmajorgrids,
ymajorgrids
]
\addplot [very thin, blue, mark=triangle*, mark size=2, mark options={rotate=180,draw=black}]
coordinates {
(1000,1)
(4641,0.9)
(21544,0.7)
(100000,0.15)
(464158,0)
(2154434,0)
(10000000,0)

};
\addplot [very thin, green!50.0!black, mark=square*, mark size=2, mark options={draw=black}]
coordinates {
(1000,0.9)
(4641,0.25)
(21544,0)
(100000,0)
(464158,0)
(2154434,0)
(10000000,0)

};
\addplot [very thin, red, mark=*, mark size=2, mark options={draw=black}]
coordinates {
(1000,1)
(4641,0.95)
(21544,1)
(100000,0.9)
(464158,0.95)
(2154434,0.75)
(10000000,0.75)

};
\addplot [very thin, color0, mark=triangle*, mark size=2, mark options={draw=black}]
coordinates {
(1000,0.95)
(4641,1)
(21544,0.95)
(100000,1)
(464158,0.75)
(2154434,0.45)
(10000000,0.45)

};
\addplot [very thin, color1, mark=triangle*, mark size=2, mark options={rotate=270,draw=black}]
coordinates {
(1000,0.9)
(4641,0.35)
(21544,0.15)
(100000,0)
(464158,0)
(2154434,0)
(10000000,0)

};
\addplot [very thin, color2, mark=triangle*, mark size=2, mark options={rotate=90,draw=black}]
coordinates {
(1000,1)
(4641,1)
(21544,0.9)
(100000,0.7)
(464158,0.4)
(2154434,0.05)
(10000000,0)

};
\addplot [very thin, black, mark=diamond*, mark size=2]
coordinates {
(1000,1)
(4641,1)
(21544,1)
(100000,1)
(464158,1)
(2154434,1)
(10000000,0.95)

};
\path [draw=black, fill opacity=0] (axis cs:1000,1)--(axis cs:10000000,1);

\path [draw=black, fill opacity=0] (axis cs:1,0)--(axis cs:1,1);

\path [draw=black, fill opacity=0] (axis cs:1000,0)--(axis cs:10000000,0);

\path [draw=black, fill opacity=0] (axis cs:0,0)--(axis cs:0,1);

\end{semilogxaxis}

\end{tikzpicture}

%% file: figures/chap5_SL_bench/time_avg.tikz
%
%
%
%
\begin{tikzpicture}

\definecolor{color1}{rgb}{0.75,0,0.75}
\definecolor{color0}{rgb}{0,0.75,0.75}
\definecolor{color2}{rgb}{0.75,0.75,0}

\begin{semilogxaxis}[
xlabel={samples},
ylabel={time [s]},
xmin=1000, xmax=10000000,
ymin=0, ymax=0.6,
axis on top,
width=\figurewidth,
height=\figureheight,
xmajorgrids,
ymajorgrids,
legend columns=2,
legend style={at={(0.03,0.74)}, anchor=south west, font=\tiny},
legend entries={{Ex. 2},{Ex. 3},{Ex. 4},{Ex. 5},{Ex. 6},{Ex. 7},{Ex. 8}}
]
\addplot [very thin, blue, mark=triangle*, mark size=2, mark options={rotate=180,draw=black}]
coordinates {
(1000,0.0318135537)
(4641,0.0289950043)
(21544,0.022471762)
(100000,0.016080717)
(464158,0.025679388)
(2154434,0.1089609881)
(10000000,0.5015200966)

};
\addplot [very thin, green!50.0!black, mark=square*, mark size=2, mark options={draw=black}]
coordinates {
(1000,0.0639878248)
(4641,0.0297312811)
(21544,0.0091067268)
(100000,0.0076644214)
(464158,0.0256326187)
(2154434,0.1088262111)
(10000000,0.5142365129)

};
\addplot [very thin, red, mark=*, mark size=2, mark options={draw=black}]
coordinates {
(1000,0.0320312895)
(4641,0.0306662861)
(21544,0.0305111005)
(100000,0.0324221458)
(464158,0.0515278907)
(2154434,0.1308113547)
(10000000,0.5321380449)

};
\addplot [very thin, color0, mark=triangle*, mark size=2, mark options={draw=black}]
coordinates {
(1000,0.0313438578)
(4641,0.0326754274)
(21544,0.0308698409)
(100000,0.0381693699)
(464158,0.0457005202)
(2154434,0.123914483)
(10000000,0.5573434299)

};
\addplot [very thin, color1, mark=triangle*, mark size=2, mark options={rotate=270,draw=black}]
coordinates {
(1000,0.029541997)
(4641,0.0149476093)
(21544,0.0062739028)
(100000,0.0063933104)
(464158,0.0296840893)
(2154434,0.109795926)
(10000000,0.4974677332)

};
\addplot [very thin, color2, mark=triangle*, mark size=2, mark options={rotate=90,draw=black}]
coordinates {
(1000,0.0326415355)
(4641,0.0305385914)
(21544,0.0305991708)
(100000,0.025541965)
(464158,0.0466390026)
(2154434,0.1157666458)
(10000000,0.5324225227)

};
\addplot [very thin, black, mark=diamond*, mark size=2]
coordinates {
(1000,0.0663556166)
(4641,0.0650094551)
(21544,0.064456473)
(100000,0.0710106154)
(464158,0.0861597717)
(2154434,0.1693928494)
(10000000,0.5609619285)

};
\path [draw=black, fill opacity=0] (axis cs:1000,1)--(axis cs:10000000,1);

\path [draw=black, fill opacity=0] (axis cs:1,0)--(axis cs:1,0.6);

\path [draw=black, fill opacity=0] (axis cs:1000,0)--(axis cs:10000000,0);

\path [draw=black, fill opacity=0] (axis cs:0,0)--(axis cs:0,0.6);

\end{semilogxaxis}

\end{tikzpicture}

%% file: figures/chap5_EM/time_avg.tikz
%
%
%
%
\begin{tikzpicture}

\begin{semilogxaxis}[
xlabel={samples},
ylabel={time [s]},
xmin=1000, xmax=10000000,
ymin=0, ymax=600,
axis on top,
width=\figurewidth,
height=\figureheight,
ytick={0,100,200,300,400,500,600},
yticklabels={0,100,200,300,400,500,600},
xmajorgrids,
ymajorgrids,
legend entries={{Ex. 2},{Ex. 3},{Ex. 4}},
legend style={at={(0.97,0.97)}, anchor=north east,font=\normalsize}
]
\addplot [very thin, blue, mark=triangle*, mark size=2, mark options={rotate=180,draw=black}]
coordinates {
(1000,9.200242136)
(3981,34.2204045405)
(15848,134.0582813655)
(63095,554.467997459)

};
\addplot [very thin, green!50.0!black, mark=square*, mark size=2, mark options={draw=black}]
coordinates {
(1000,5.787624443)
(3981,37.1887944675)
(15848,110.6678162175)
(63095,462.151227118)

};
\addplot [very thin, red, mark=*, mark size=2, mark options={draw=black}]
coordinates {
(1000,8.924926532)
(3981,35.4149606845)
(15848,137.334330069)
(63095,567.8513991515)

};
\path [draw=black, fill opacity=0] (axis cs:1000,1)--(axis cs:10000000,1);

\path [draw=black, fill opacity=0] (axis cs:1,0)--(axis cs:1,600);

\path [draw=black, fill opacity=0] (axis cs:1000,0)--(axis cs:10000000,0);

\path [draw=black, fill opacity=0] (axis cs:0,0)--(axis cs:0,600);

\end{semilogxaxis}

\end{tikzpicture}

%% file: figures/chap5_EM/O_dev.tikz
%
%
%
%
\begin{tikzpicture}

\begin{loglogaxis}[
xlabel={samples},
ylabel={error of $O$-estimate},
xmin=1000, xmax=100000,
ymin=1e-07, ymax=10,
axis on top,
width=\figurewidth,
height=\figureheight,
xmajorgrids,
ymajorgrids,
legend entries={{EM Ex. 2},{EM Ex. 3},{EM Ex. 4}},
legend style={at={(0.03,0.03)}, anchor=south west}
]
\addplot [very thin, blue, mark=triangle*, mark size=2, mark options={rotate=180,draw=black}]
coordinates {
(1000,0.1120377358)
(3981,0.1516811919)
(15848,0.083195909)
(63095,0.1030100693)

};
\addplot [very thin, green!50.0!black, mark=square*, mark size=2, mark options={draw=black}]
coordinates {
(1000,0.02877484)
(3981,0.0204081702)
(15848,0.0022287768)
(63095,0.0205290102)

};
\addplot [very thin, red, mark=*, mark size=2, mark options={draw=black}]
coordinates {
(1000,0.047638448)
(3981,0.0402909564)
(15848,0.0654355337)
(63095,0.0449834755)

};

\end{loglogaxis}

\end{tikzpicture}

%% file: figures/chap5_EM/T_dev.tikz
%
%
%
%
\begin{tikzpicture}

\begin{loglogaxis}[
xlabel={samples},
ylabel={error of $T$-estimate},
xmin=1000, xmax=100000,
ymin=1e-06, ymax=100,
axis on top,
width=\figurewidth,
height=\figureheight,
xmajorgrids,
ymajorgrids
]
\addplot [very thin, blue, mark=triangle*, mark size=2, mark options={rotate=180,draw=black}]
coordinates {
(1000,0.0642828175)
(3981,0.016898522)
(15848,0.0226031794)
(63095,0.0034489199)

};
\addplot [very thin, green!50.0!black, mark=square*, mark size=2, mark options={draw=black}]
coordinates {
(1000,0.0031888974)
(3981,0.0177881626)
(15848,0.02770989475)
(63095,0.071885496)

};
\addplot [very thin, red, mark=*, mark size=2, mark options={draw=black}]
coordinates {
(1000,0.01826411)
(3981,0.016939758)
(15848,0.0215668326)
(63095,0.045913576)

};

\end{loglogaxis}

\end{tikzpicture}

%% file: figures/EM_with_true_and_SL/O_dev.tikz
%
%
%
%
\begin{tikzpicture}

\definecolor{color1}{rgb}{0.75,0,0.75}
\definecolor{color0}{rgb}{0,0.75,0.75}
\definecolor{color2}{rgb}{0.75,0.75,0}

\begin{axis}[
xlabel={samples},
ylabel={error of $O$-estimate},
xmin=1000, xmax=10000000,
ymin=1e-07, ymax=10,
xmode=log,
ymode=log,
axis on top,
width=\figurewidth,
height=\figureheight,
xmajorgrids,
ymajorgrids,
legend columns=2,
legend style={at={(0.03,0.03)}, anchor=south west, font=\scriptsize},
legend entries={{EM Ex. 2},{EM Ex. 3},{EM Ex. 4},{SL Ex. 2},{SL Ex. 3},{SL Ex. 4}}
]
\addplot [very thin, blue, mark=triangle*, mark size=2, mark options={rotate=180,draw=black}]
coordinates {
(1000,0.119371683)
(3981.00000000001,0.039024098)
(15848,0.0265514869999999)
(63095.0000000001,0.000596680599999999)

};
\addplot [very thin, green!50.0!black, mark=square*, mark size=2, mark options={draw=black}]
coordinates {
(1000,0.0053888685)
(3981.00000000001,0.0001646037)
(15848,5.91168e-05)
(63095.0000000001,1.46268e-05)

};
\addplot [very thin, red, mark=*, mark size=2, mark options={draw=black}]
coordinates {
(1000,0.0660588285999999)
(3981.00000000001,0.00838511569999999)
(15848,0.0013666179)
(63095.0000000001,0.000443270599999999)

};
\addplot [very thin, blue, dashed, mark=triangle*, mark size=2, mark
options={rotate=180,solid,draw=black}]
coordinates {
(1000,0.420686759999999)
(4641,4.9121650574)
(21544,1.1980582023)
(100000,0.00759266339999998)
(464158,0.0012276053)
(2154433.99999999,0.0002602418)
(10000000,4.55481e-05)

};
\addplot [very thin, green!50.0!black, dashed, mark=square*, mark size=2, mark
options={draw=black,solid}]
coordinates {
(1000,0.0250202536)
(4641,0.0064876399)
(21544,0.0001257871)
(100000,2.96942e-05)
(464158,5.8149e-06)
(2154433.99999999,1.3195e-06)
(10000000,3.03e-07)

};
\addplot [very thin, red, dashed, mark=*, mark size=2, mark
options={draw=black,solid}]
coordinates {
(1000,1.8094939353)
(4641,0.511298839099999)
(21544,0.814743959299998)
(100000,0.4045767371)
(464158,0.2482228942)
(2154433.99999999,0.151350762)
(10000000,0.2749241359)

};

\end{axis}

\end{tikzpicture}

%% file: figures/EM_with_true_and_SL/T_dev.tikz
%
%
%
%
\begin{tikzpicture}

\definecolor{color1}{rgb}{0.75,0,0.75}
\definecolor{color0}{rgb}{0,0.75,0.75}
\definecolor{color2}{rgb}{0.75,0.75,0}

\begin{axis}[
xlabel={samples},
ylabel={error of $T$-estimate},
xmin=1000, xmax=10000000,
ymin=1e-06, ymax=100,
xmode=log,
ymode=log,
axis on top,
width=\figurewidth,
height=\figureheight,
xmajorgrids,
ymajorgrids
]
\addplot [very thin, blue, mark=triangle*, mark size=2, mark options={rotate=180,draw=black}]
coordinates {
(1000,0.0206995173)
(3981.00000000001,0.00707860419999999)
(15848,0.0017812776)
(63095.0000000001,0.0006235086)

};
\addplot [very thin, green!50.0!black, mark=square*, mark size=2, mark options={draw=black}]
coordinates {
(1000,0.002285246)
(3981.00000000001,0.0005652431)
(15848,0.0002339459)
(63095.0000000001,4.64658e-05)

};
\addplot [very thin, red, mark=*, mark size=2, mark options={draw=black}]
coordinates {
(1000,0.0278542926)
(3981.00000000001,0.011649024)
(15848,0.0019828211)
(63095.0000000001,0.000689033299999999)

};
\addplot [very thin, blue, dashed, mark=triangle*, mark size=2, mark
options={rotate=180,draw=black,solid}]
coordinates {
(1000,2.044555523)
(4641,0.3422954473)
(21544,0.0713145181999999)
(100000,0.0046401567)
(464158,0.0013853535)
(2154433.99999999,0.0002382684)
(10000000,4.828e-05)

};
\addplot [very thin, green!50.0!black, dashed, mark=square*, mark size=2, mark
options={draw=black,solid}]
coordinates {
(1000,0.1412195755)
(4641,0.0039895532)
(21544,0.0005980195)
(100000,0.0001330955)
(464158,2.04241e-05)
(2154433.99999999,7.52259999999999e-06)
(10000000,1.3918e-06)

};
\addplot [very thin, red, dashed, mark=*, mark size=2, mark
options={draw=black,solid}]
coordinates {
(1000,10.9115376712)
(4641,0.6947624578)
(21544,1.121573524)
(100000,1.8971174293)
(464158,0.347450855399999)
(2154433.99999999,0.1636030512)
(10000000,0.1122621116)

};

\end{axis}

\end{tikzpicture}